\documentclass[10pt,twocolumn]{article}
\usepackage{authblk}

\usepackage[T1]{fontenc}
\usepackage[utf8]{inputenc}
\usepackage{authblk}
\usepackage{tikz}
\usepackage{float}
\usepackage{hyperref}
\usepackage{multicol}
\usepackage[]{algorithm2e}
\usepackage{cuted}
\usepackage{blindtext}
\usepackage[english]{babel}
\usepackage{blindtext}
\usepackage{multicol}
\usepackage{etoolbox}
\usepackage{booktabs}
\usepackage{lipsum}
\usepackage{parskip}
\usepackage{here}
\usepackage[font=small,labelfont=bf]{caption}

\usetikzlibrary{shapes}

\usetikzlibrary{shapes.gates.logic.US,trees,positioning,arrows}

\title{Genetic Network Architecture Search}

\author[1]{Hai Victor Habi\thanks{haivictorh@mail.tau.ac.il}}
\author[2]{Gil Rafalovich}
\affil[1,2]{The School of Electrical Engineering \protect\\ Tel Aviv University}

\begin{document}

\maketitle
\thispagestyle{empty}
\pagestyle{empty}
\raggedbottom
\begin{abstract}
We propose a method\footnote{This project was done during Deep Learning Course (0510725501) in Tel Aviv University in Dec, 2018 .}  for learning the neural network architecture that based on Genetic Algorithm (GA). Our approach uses a genetic algorithm integrated with standard Stochastic Gradient Descent(SGD) which allows the sharing of weights across all architecture solutions.  The method uses GA to design a sub-graph of Convolution cell which maximizes the accuracy on the validation-set. Through experiments, we demonstrate this method's performance on both CIFAR10 and CIFAR100 dataset with an accuracy of $96\%$ and $80.1\%$. The code and result of this work available in GitHub:\href{https://github.com/haihabi/GeneticNAS}{https://github.com/haihabi/GeneticNAS}.
\end{abstract}
\section{Introduction}
Neural networks produced outstanding results in many fields such as computer vision and language models with a wide variety of tasks such as classification, object detection, segmentation, etc. The design of neural network architectures for a specific task or dataset requires a large amount of knowledge and experience. The use of Network Architectures Search (NAS) has shown a lot of success in the design of architectures for image classification
and language models \cite{zoph2016neural}\cite{zoph2018learning}\cite{cai2018efficient}\cite{liu2018progressive}. In NAS, a recurrent neural network (RNN) controller is trained to generate a candidate network architecture (e.g. child model) which is then trained to converge. Once the training is complete, a measure of performance is taken using the trained network architecture on the desired task or dataset. The controller uses the performance measurement as a signal for finding a better architecture, this process is repeated over many iterations, which are computationally expensive. In recent year several studies  \cite{liu2018darts}\cite{pham2018efficient} used the idea of weight sharing across all child models to reduce much of the computational effort of the search, where weight sharing has shown prominent result using refinement based method \cite{pham2018efficient} and gradient based method\cite{liu2018darts}. 
\\\\
In this work, we present a genetic network architecture search (GeneticNAS) a method based on genetic algorithm which optimized the network architecture for finding a good architecture on the validation-set. Our work utilizes the search space, defined by \cite{pham2018efficient} which presents a convolution cell as a directed acyclic graph (DAG) with a fixed length list of integers that are used to describe the DAG structure. Furthermore, the use of evolution based method for NAS has been used by \cite{8237416}\cite{liu2017hierarchical}\cite{real2018regularized}, but without weight sharing . In this work we show for the first time a weights sharing with evolution base for convolution architecture search.   
\\\\
This work, organized as follows: In Section \ref{related}, we provide an overview of the related works in the field. In Section \ref{data}, we provide full details on the data set used in this study. In
Section \ref{method}, we describe the network architecture search method and the search space. Section \ref{experiments} describes the experiments and the results, while Section \ref{discussion} provides conclusions and a discussion.
\section{Related Work}\label{related}
NAS has shown promising results in image classification and language modeling tasks but with highly computationally expensive. In the last year multiple studies have shown the use of weight sharing a cross candidate model that reduces the need for training each child from the beginning thus removing most of the computationally expenses. The works that utilize weight sharing is divided into two methods. The first is efficient neural architecture search via parameter sharing (ENAS) \cite{pham2018efficient}. This is base on reinforcement learning which use a RNN to generate candidate architecture. The second method:is a differentiable architecture search (DARTS) \cite{liu2018darts} which is base on gradients where each connection has a probability function that is updated by the gradient. 
\\\\
A genetic algorithm is a search method which is base on natural selection and genetics, GAs are built on four key components: selection, cross-over, mutation and replacement. All the operations are performed on an encoding of the search space called 'chromosome', where each value in a chromosome is called a 'genotype' (gen) and a specific solution is referred on to 'individual'. An other key concept of GAs is a 'population' which is used for generating new candidate solutions. In each iteration a new generation is created using the three step selection, cross-over and mutation process. The new generation is then inserted into the population using the replacement step. The algorithm starts with a random population which is evaluated at the onset. There are several works (e.g. \cite{liu2017hierarchical} \cite{real2018regularized}) which utilize the evolution method for neural architecture searching but without weight sharing.
\raggedbottom
\section{Data}\label{data}
\begin{figure}[H]
	\includegraphics[scale=0.4]{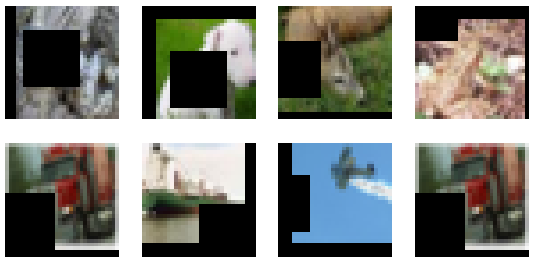}
	\centering
	\caption{Image after data augmentation with CutOut}
	\label{fig:cutout}
\end{figure}

The data used in the experiment is CIFAR10, and CIFAR100 datasets\cite{krizhevsky2009learning} which contain 50000 training images and 10000 validation images. In both datasets we use the same preprocessing and augmentation procedures that have been used in a previous work \cite{He_2016_CVPR}. We preprocess the data by normalizing all images. Additionally, we augmented the data by padding the image with 4 pixels on each side, and a 32x32 crop is randomly sampled from the padded image. We also use random horizontal flips on this 32x32 cropped image. Finally we apply CutOut \cite{devries2017improved} on the flipped image. An example image post augmentation is shown in Figure \ref{fig:cutout}. 
\section{Method}\label{method}
The main idea of GeneticNAS is the combination of ENAS with an evolution search. Additionally we have used the ENAS search space that represents a network cell as DAG where each node represents an operation and the edges represent a tensors. The key property of ENAS - weight sharing across all child networks is applied in GeneticNAS as well. 

\raggedbottom
\subsection{CNN}
\subsubsection{Search Space}
\begin{figure}[H]
	\includegraphics[scale=0.8]{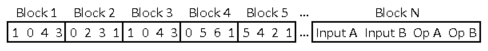}
	\centering
	\caption{An integer list representing a cell type}
	\label{fig:ind_list}
\end{figure}

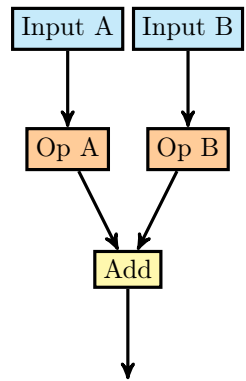
\begin{figure}[H]
    \centering
    \begin{tikzpicture}[
    and/.style={and gate US,thick,draw,fill=red!60,rotate=90,
		anchor=east,xshift=-1mm},
    or/.style={or gate US,thick,draw,fill=blue!60,rotate=90,
		anchor=east,xshift=-1mm},
    be/.style={circle,thick,draw,fill=green!60,anchor=north,
		minimum width=0.7cm},
    tr/.style={buffer gate US,thick,draw,fill=purple!60,rotate=90,
		anchor=east,minimum width=0.8cm},
    label distance=3mm,
    every label/.style={blue},
    event/.style={rectangle,thick,draw,fill=yellow!20,text width=2cm,
		text centered,font=\sffamily,anchor=north},
    edge from parent/.style={very thick,draw=black!70},
    edge from parent path={(\tikzparentnode.south) -- ++(0,-1.05cm)
			-| (\tikzchildnode.north)},
    level 1/.style={sibling distance=7cm,level distance=1.4cm,
			growth parent anchor=south,nodes=event},
    level 2/.style={sibling distance=7cm},
    level 3/.style={sibling distance=6cm},
    level 4/.style={sibling distance=3cm}
    ]


   \begin{scope}[very thick,
		node distance=1.6cm,on grid,>=stealth',
		block/.style={rectangle,draw,fill=cyan!20},
		comp/.style={rectangle,draw,fill=orange!40},
		comp_add/.style={rectangle,draw,fill=yellow!40}]
   \node [] (re)					{};
   \node [comp_add]	 (cb)	[above=of re]			{Add}  edge [->] (re);
   \node [comp]	 (ca1)	[above=of cb, xshift=-0.8cm]	{Op A} edge [->] (cb);
   \node [comp]	 (ca2)	[right=of ca1]			{Op B} edge [->] (cb);
   \node [block] (s1)	[above=of ca1]		{Input A} edge [->] (ca1);
   \node [block] (s2)	[right=of s1]		{Input B} edge [->] (ca2);
   \end{scope}
\end{tikzpicture}
    \caption{The block structure with two inputs, two operation and merge operator}
    \label{fig:block}
\end{figure}

The convolution search space is based on the idea that the same cell is repeated several times, which is common practice in the design of a neural network (e.g \cite{He_2016_CVPR} \cite{xie2017aggregated} \cite{howard2017mobilenets} \cite{szegedy2017inception} \cite{sandler2018mobilenetv2} etc). The cell is inspired by \cite{pham2018efficient} which uses the DAG representation that include $N_b$ blocks, where each block contains two inputs, two operators and ends with a merge operation. In this work, we select add as a merge operation. The block structure is shown in Figure \ref{fig:block}. 
Each block has four integer representations for the operation and connectivity to the other blocks in the cell. The first two integers $i_A,i_B$ represent the input of the block, where the second two integers $j_A,j_B$ represent the operation type of each input. Moreover, the input integer describes each possible connection to the block including the input of the DAG - where each block can only receive input from the block below. For example, the second block can receive input from blocks 0 ,1 and the DAG input. The DAG is represented by a list of integers as shown in Figure \ref{fig:ind_list}. The final DAG operation is a concatenation of all blocks that are not used as input to any other block. The concatenation operation is followed by a 1x1 convolution that align the number of output channels to the number of channels in the DAG operation. Furthermore all operations in the DAG have the same number of channels.
\begin{equation}
    \label{eq:op}
    h_{n}=OP_{i_A,j_A}^{A}(h_{i_A})+OP_{i_B,j_B}^{B}(h_{i_B})
\end{equation}
The operations selected for this work are same as the ones selected by \cite{pham2018efficient} and are presented in the list below:
\begin{itemize}
    \item identity
    \item 3x3 max pooling
    \item 3x3 average pooling
    \item 3x3 depth wise-separable convolution
    \item 5x5 depth wise-separable convolution
\end{itemize}
The pooling operations start with ReLU non-linear then followed by an average or max pooling. The operation output has the same shape as the input using a zero padding and stride one.
The depth wise-separable convolution has the following structure which is ReLU, depth-wise convolution, batch-normalization\cite{ioffe2015batch} (BN) then ReLU, point-wise convolution and BN. This structure is repeated twice in-order to achieve a greater depth. Moreover, all convolution operations output have the same shape as the input using a zero padding and stride one. Similar to ENAS we have introduced weight sharing between models where each input has a separate set of weight per operation as shown by Equation \ref{eq:op} which describes the output of Block $n$.\\

\begin{figure}[H]
	\includegraphics[scale=0.3]{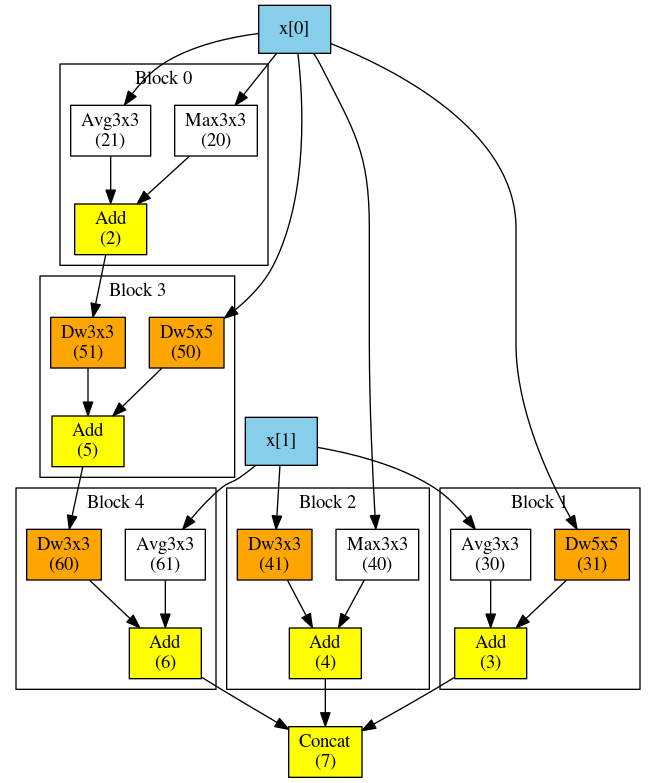}
	\centering
	\caption{An example DAG with 5 block}
	\label{fig:example_dag}
\end{figure} 
Figure \ref{fig:example_dag} is an example of DAG with five blocks, where, for example, block 2's integer list is the following $[1,0,3,1]$.

Our search space includes an exponential number of convolution cells. Specifically, if the convolution cell has $N_b$ blocks and we allow $N_{op}$ operators, then the search space has $N_{op}^{N_b}\cdot(N_b!)^2$ configurations. In our experiments, $N_b=5$ and $N_{op}=5$, which gives there are approximately $10^{12}$ convolution cells in our search space.
\subsubsection{Network Architecture}
\begin{figure}[h]
    \centering
    \begin{tikzpicture}[
    and/.style={and gate US,thick,draw,fill=red!60,rotate=90,
		anchor=east,xshift=-1mm},
    or/.style={or gate US,thick,draw,fill=blue!60,rotate=90,
		anchor=east,xshift=-1mm},
    be/.style={circle,thick,draw,fill=green!60,anchor=north,
		minimum width=0.7cm},
    tr/.style={buffer gate US,thick,draw,fill=purple!60,rotate=90,
		anchor=east,minimum width=0.8cm},
    label distance=3mm,
    every label/.style={blue},
    event/.style={rectangle,thick,draw,fill=yellow!20,text width=2cm,
		text centered,font=\sffamily,anchor=north},
    edge from parent/.style={very thick,draw=black!70},
    edge from parent path={(\tikzparentnode.south) -- ++(0,-1.05cm)
			-| (\tikzchildnode.north)},
    level 1/.style={sibling distance=7cm,level distance=1.4cm,
			growth parent anchor=south,nodes=event},
    level 2/.style={sibling distance=7cm},
    level 3/.style={sibling distance=6cm},
    level 4/.style={sibling distance=3cm}
    ]



   \begin{scope}[very thick,
		node distance=1.8cm,on grid,>=stealth',
		input/.style={circle,draw,fill=white!20,minimum size=1.2cm},
		block/.style={rectangle,draw,fill=cyan!20},
		output/.style={circle,draw,fill=blue!40},
		add/.style={circle,draw,fill=yellow!40,minimum size=1.2cm}]
   \node [input] (input_0)	[]		{$h_n$} ;
   \node [input] (input_1)	[right=of input_0]		{$h_{n-1}$} ;
   \node [cloud, draw,cloud puffs=10,cloud puff arc=120, aspect=2, inner ysep=1em,above=of input_1,xshift=-0.8cm,minimum size=2cm](base_cell) {DAG} edge [<-] (input_1) edge [<-] (input_0);
   \node [block] (se)	[above=of base_cell]		{SE Block} edge [<-] (base_cell);
   \node [add] (add)	[above=of se]		{$+$} edge [<-] (se) edge [<-,bend right=60] (input_0);
    \node [output] (se)	[above=of add]		{$h_{n+1}$} edge [<-] (add);
   \end{scope} 
   \begin{scope}[xshift=-4cm,very thick,
		node distance=1.0cm,on grid,>=stealth',
		block/.style={rectangle,draw,fill=cyan!20},
		block_image/.style={rectangle,draw,fill=green!20},
		block_reduce/.style={rectangle,draw,fill=red!20},
		block_input/.style={rectangle,draw,fill=orange!20} ]
	 \node [block_image] (input_b)	[]		{Input} ;
	 \node [block_input] (cell1)	[above=of input_b]		{Input Cell} edge [<-] (input_b);
	 \node [block] (cell2)	[above=of cell1]		{Normal Cell}edge [<-] (cell1);
	 \node [block_reduce] (cell3)	[above=of cell2]		{Reduction Cell} edge [<-] (cell2);
	 \node [block] (cell4)	[above=of cell3]		{Normal Cell}edge [<-] (cell3);
	 \node [block_reduce] (cell5)	[above=of cell4]		{Reduction Cell}edge [<-] (cell4);
	 \node [block] (cell6)	[above=of cell5]		{Normal Cell}edge [<-] (cell5);
	 \node[right=of cell2,xshift=0.9cm]  {x $N_{cell}$ -1};
	 \node[right=of cell4,xshift=0.8cm]  {x $N_{cell}$};
	 \node[right=of cell6,xshift=0.8cm]  {x $N_{cell}$};

   \end{scope}
\end{tikzpicture}
    \caption{Left Figure present the network structure with three cell types, Right Figure present the cell structure with residual connection and Squeeze and Excitation block  }
    \label{fig:cell}
\end{figure}
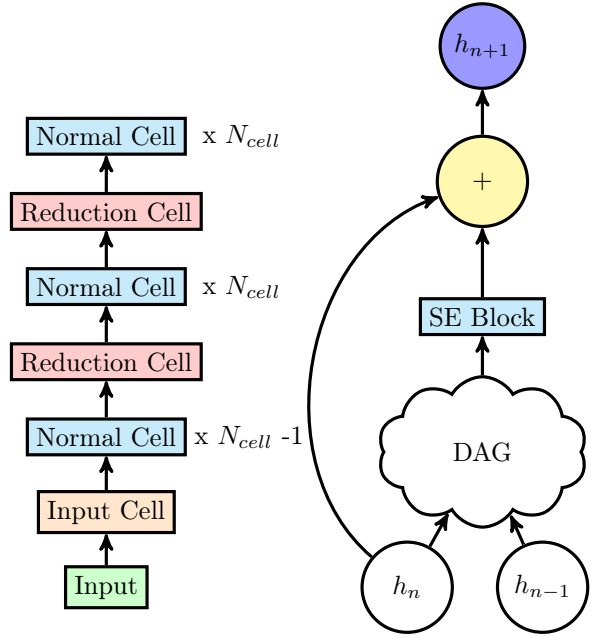

The network architecture is built with three type of cells. An input cell, a Normal Cell and a Reduction Cell.The network structure begins with an Input Cell, followed by 3 sets of cells. Each set built using $N_{cell}$ or $N_{cell}-1$ normal cells, that followed by a reduction cell aside from the last set. The network structure is presented in Figure \ref{fig:cell}. The Normal Cell and Reduction Cell have two inputs, $h_n$ from previous cell output and $h_{n-1}$ from previous cell input. The current cell has a single output $h_{n+1}$. In addition all cell types have a Squeeze and Excitation block \cite{hu2018squeeze} with residual connection \cite{He_2016_CVPR} as presented in Figure \ref{fig:cell}.
The Input Cell different due to its use as the first cell of the network where there is no bypass connection from the previous cell. Furthermore, the Input Cell performs a 3x3 convolution on his input for expanding the number of channels from RGB to $N_c$. The Reduction Cell used after every dimensions reduction. The dimensions reduction uses average pooling with stride two that followed by a 1x1 convolution which expanding the number of a channel by a factor of two. 
\subsection{Architecture Search}
The architecture search is built on a GA integrated into stochastic gradient descent (SGD) that utilizes weight sharing across all network architecture, which described in section \ref{ss:int}. For GA we will define all the components of the search (Selection,  Cross-over, Mutation  and  Replacement) that described in the following section. 
\subsubsection{Selection}
Selection use the fitness $f_i$ of each individual in the current population in-order to calculate the probability of an individual, normalized by the sum of all individual fitness.
\begin{equation}
    \label{eq:selection}
    p_i=\frac{f_i}{\sum_i f_i}
\end{equation}
 We than randomly select $N_p$ individuals from the current population using the probability calculated by Equation \ref{eq:selection}.
\subsubsection{Cross-over}
\begin{figure}[H]
	\includegraphics[scale=0.8]{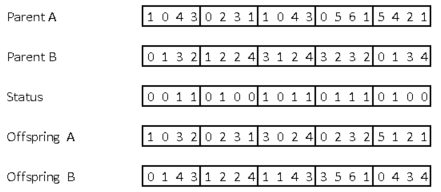}
	\centering
	\caption{Uniform crossover method}
	\label{fig:block_u}
\end{figure}

The cross-over operation takes two individuals that called 'parents', then produce a two new individuals that called 'offsprings'. In this work we have used two types of cross-over methods. The first method is a unifrom cross-over, commonly used with GA, where the second cross-over is a block cross-over which is specifically designed for the representation of network architecture.
\paragraph{Uniform Cross-over}
The uniform cross-over is a common cross-over operation used in GAs. It begins by generating a random binary vector in the size of the chromosome. The second parent gene swap with the first parent gene if the binary vector equal to one, otherwise the gene is taken from the first parent. The first offspring produced by this process, while the second offspring produced by the same process after applying not on the binary vector. An example is presented in Figure \ref{fig:block_u}. 
\paragraph{Block Cross-over}
Block cross-over utilizes the fact that a cell type is built with $N_b$ block. We begin by defining the cross-over blocks by generating a random binary status vector with size $N_b$. The process for producing the first offspring is as follows: If the status equals one, the first parent swaps its block with the second parent, otherwise the block is taken from the first parent. The second offspring is produced by the same process after applying not on the binary vector.
\begin{figure}[H]
	\includegraphics[scale=0.8]{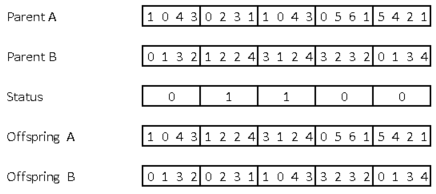}
	\centering
	\caption{Block crossover method}
	\label{fig:block_cv}
\end{figure}
\subsubsection{Mutation}
Mutation modifies the individual representation randomly. For each generation a random value, zero or one, is generated, where the probability for one is $p_m$. This random value indicates which gene is to be mutated. If a gene is selected to be mutated then we randomly add $+1$ or $-1$ to its representation with equal probability.
\begin{figure}[H]
	\includegraphics[scale=0.7]{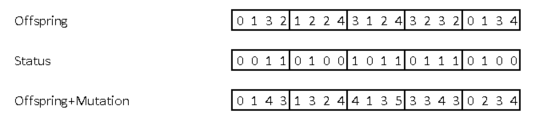}
	\centering
	\caption{Offspring mutation example}
	\label{fig:mutation_example}
\end{figure}
\subsubsection{Replacement}
One of the key properties of GA is a population, where in each iteration a set of a new generation of individuals is inserted in to the population and a current set of the individuals, (within the current population) is removed.
This ensures the same population size of $N_p$. There are several known replacement methods for GA as shown in \cite{burke2005search}. We have chosen a steady-state-no-duplicates method. This method ensures that the population is comprised of the best $N_p$ individuals derived from the current population and the new generation.
\subsubsection{Integration} \label{ss:int}
We integrated the GA search with a Stochastic Gradient Descent (SGD) algorithm, for efficient architecture search. Weight sharing was utilized which reduces the need to train all child networks from the beginning. The weight sharing requires the model coefficient to be adaptable to network architecture change. This requires an update of all the weights. In-order to achieve this update, in each batch we sample a different architecture from the current population. The sampling process contains three stages:randomly selecting two individuals from the current population, cross-over followed by a mutation that applied to the first offspring and returned as a result. The sampling process is presented by Algorithm \ref{alg:smaple}.
\begin{algorithm}[ht]
 \KwData{current population $x_p$}
 \KwResult{sampled individual $x_s$}
 $x_0,x_1=sampling(x_p)$\; 
 $x_0,x_1=crossover(x_0,x_1)$\;
 $x_s=mutation(x_0)$\;
 \caption{GeneticNAS sampling process}
 \label{alg:smaple}
\end{algorithm}
After the training process of a single epoch is complete, a new generation is created from the current population using the following three stages: selection, cross-over and mutation. Each individual is then evaluated on a subset of the validation set of size $N_v$. Once all individuals are evaluated, the current population is  then update using the replacement method. The evaluation of a subset of the validation reduces the computational cost of the fitness at the cost of the accuracy which allows us for more GA updates per epoch. The GeneticNAS training process is presented in Algorithm \ref{alg:gnas}.
\begin{algorithm}[h]
 \KwData{dataset}
 \KwResult{network architecture}
 initialization weights\;
 \For{each epoch}{
  \For{each batch}{
    sample random child network from population\;
    forward pass\;
    backward pass\;
  }
  generate next generation from population\;
  \For{each child in next generation}{
    set child architecture\;
    evaluated architecture accuracy\;
  }
  update population using replacement\;
 }
 \caption{GeneticNAS algorithm integrated}
 \label{alg:gnas}
\end{algorithm}
\section{Experiments}\label{experiments}
In this section, we describe a set of experiments performed with GeneticNAS. The experiments begin with an ablation study of genetic search methods using CIFAR10 dataset. We present the performance of GeneticNAS on CIFAR10 and CIFAR100 datasets, begin with an architecture search, following a final training stage on the search solution. All the experiments are performed with a single NVIDIA GTX 1080Ti graphics processing unit (GPU). In all experiments we used the following learning schedule: Beginning from 0.1, which is divided by 10 at $0.5*N_e$ and divided by 10 again at $0.75*N_e$. We have also applied weight decay of 0.0001 with dropout \cite{srivastava2014dropout} with probability of 0.2. The batch size is set to 128 images per gradient update. 
\subsection{Ablation study}
The ablation study was performed on CIFAR10 dataset with a small model of  $N_{cell}=2$. The number of channels of the first cell is $N_c=20$. The optimizer used is SGD with Nesterov momentum \cite{sutskever2013importance} which is set to $0.9$. The training is performed over 310 epochs. The generation size is set to $N_g=20$ where each individual is evaluated over 1000 images that are randomly selected from the validation set. The number of blocks used to construct the convolution cell is $N_b=5$.
\begin{figure}[H]
	\includegraphics[scale=0.5]{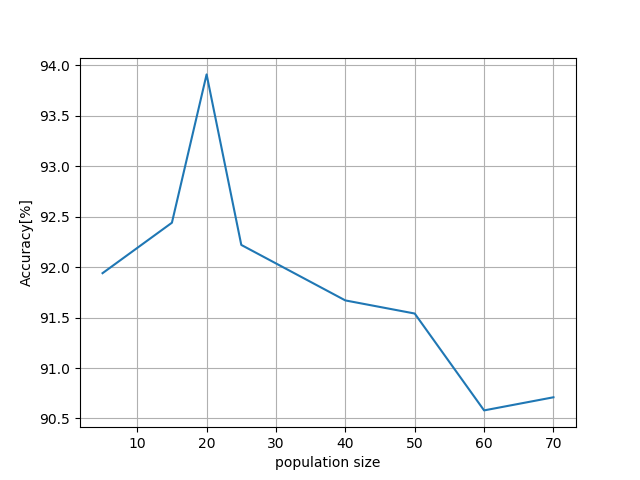}
	\centering
	\caption{Final search result over different population size}
	\label{fig:pop_size}
\end{figure}
Figure \ref{fig:pop_size} shows the result of a search with different population sizes. A higher population leads to poor performance which is caused by high variability of population. However, low population size also leads to poor performance which reduces the ability of GA to explore a new solution.
\begin{table}[H]
	\centering
	\caption{Compare search result with Uniform vs Block Cross-over}
	\label{tab:table1}
	\begin{tabular}{cc}
		\toprule
		Cross-over method & Accuracy\\
		\midrule
		Uniform & $91.8 \%$   \\
		\midrule
		Block & 93.90\%  \\
		\bottomrule
	\end{tabular}
\end{table}
Table \ref{tab:table1} presents a comparison between the uniform Cross-over and block Cross-over accuracy. The results indicates that using the block structure of the cell has increase in accuracy benefit of $2\%$. This result can be explained by the fact that block cross-over has less variability of the network architecture which allows the search to exploit the search space more efficiently as well as to utilize the structure of the DAG representation.

\begin{figure}[H]
	\includegraphics[scale=0.5]{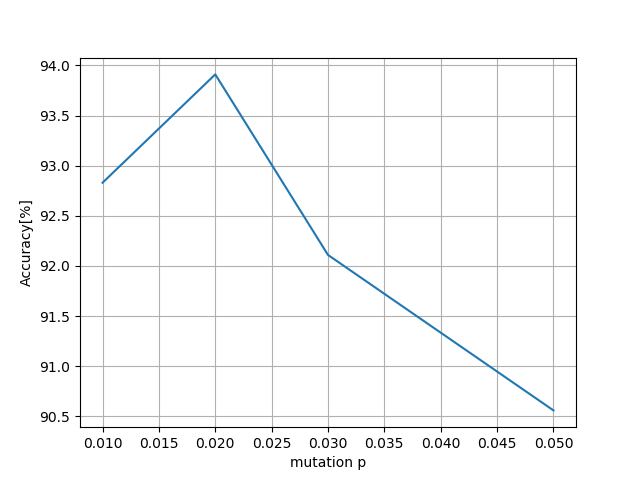}
	\centering
	\caption{Final search result over different mutation probability}
	\label{fig:mutation}
\end{figure}
Figure \ref{fig:mutation} shows the result of a search with different mutation probabilities $p_m$. This shows the trade of between exploration and exploitation, where a high mutation probability leads to poor performance caused by high variability of the network architecture and similarly for a low mutation probability which reduce the ability of GA to explore for better solutions. Both Figures \ref{fig:pop_size}, \ref{fig:mutation} and Table \ref{tab:table1} present the exploitation vs exploration trade-off. This not only affects the genetic search but also thet SGD optimization due to the fact that for each gradient update we sample a different architecture from the current population. The sampling of different architectures during training acts as a regularization method, that resembles \cite{gastaldi2017shake} and drop-path \cite{larsson2016fractalnet}. The regularization is parameterized by the genetic search. 
\subsection{CIFAR10 Result}
\begin{figure}[H]
	\includegraphics[scale=0.3]{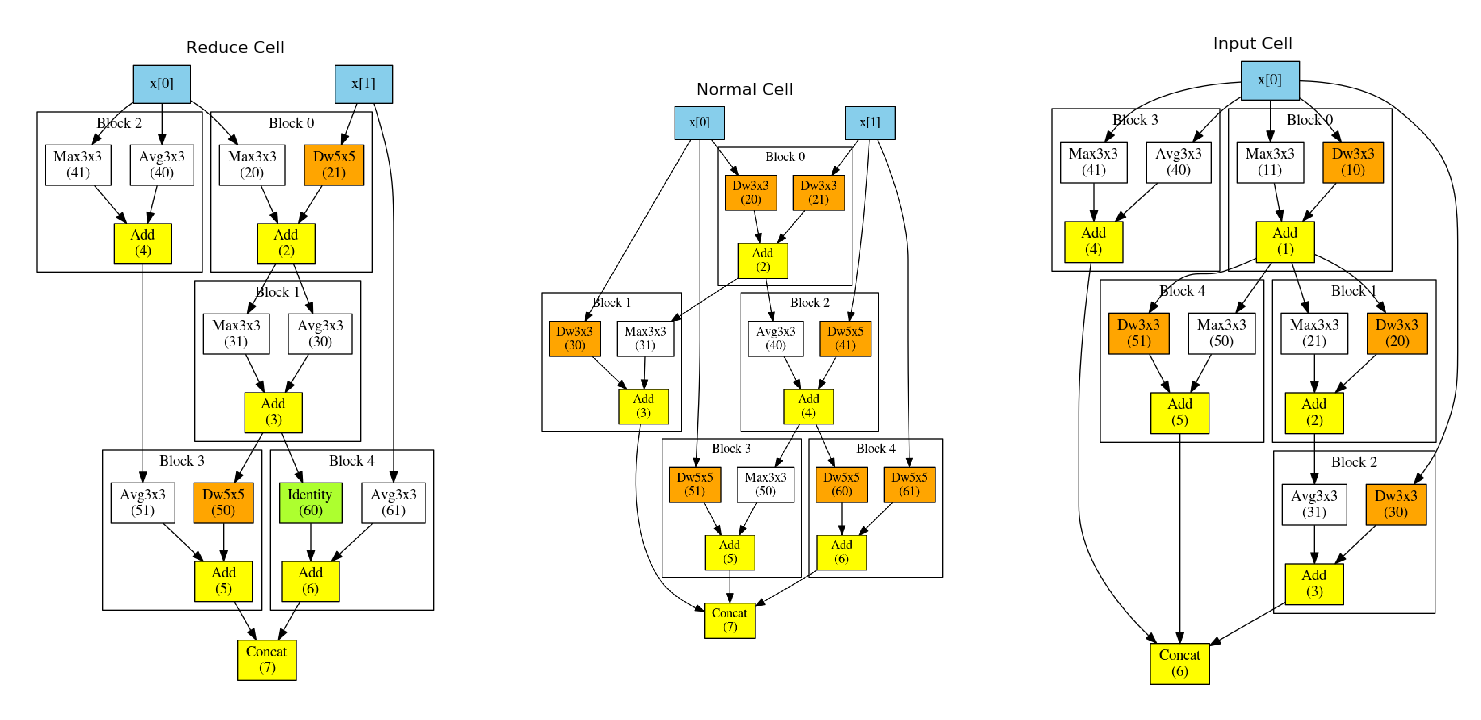}
	\centering
	\caption{Reduce Cell, Normal Cell and Input Cell Search Result on CIFAR10 dataset}
	\label{fig:search_res_cifar10}
\end{figure}

We start with an architecture search for the CIFAR10 dataset using GeneticNAS. The hyper-parameters selected are base on the ablation study, where the mutation probability $p_m=0.02$ and population size of 20. The rest of the hyper-parameters are the same as in the ablation study. The resulting architecture is presented in Figure \ref{fig:search_res_cifar10} which show the all three cell results.
\begin{figure}[H]
	\includegraphics[scale=0.4]{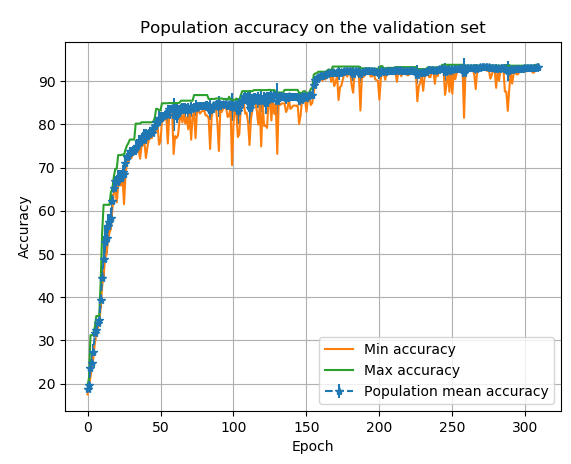}
	\centering
	\caption{The population mean accuracy during the search on CIFAR10 dataset}
	\label{fig:pop_cifar10}
\end{figure}
Figure \ref{fig:pop_cifar10} present the population mean accuracy, max accuracy, min accuracy and an error bar that indicates the standard deviation of the population accuracy on the subset of the validation-set. 
\begin{figure}[H]
	\includegraphics[scale=0.4]{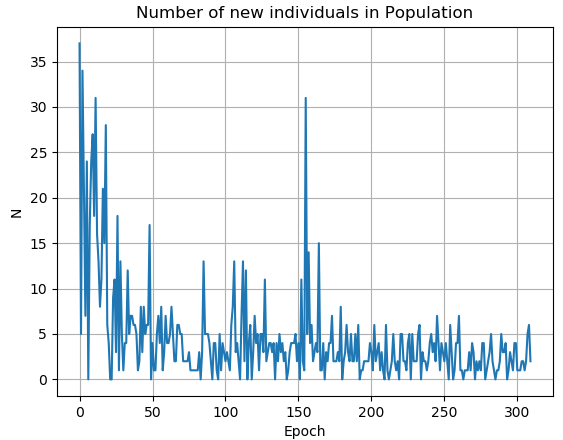}
	\centering
	\caption{Number of update to the population during training on CIFAR10 dataset}
	\label{fig:changes_cifar10}
\end{figure}
Figure \ref{fig:changes_cifar10}  presents the number of new individuals inserted into the population at the end of each epoch. This indicates the ability of the search method to generate a new and better solutions as shown in Figure \ref{fig:changes_cifar10}, where in most of the epochs, the GA population has an update. Once we have found an architecture we run a final training stage which uses the search's best result. Training is started from scratch and trained for 630 epochs. The model in the final stage has $N_{cell}=4$ cells and the number of channels is increased to $N_c=48$. For regularization we apply drop-path with a drop probability of $0.9$. The result of the final training is presented in Table \ref{tab:cifar10_acc} which compares the result on CIFAR10 on a set of different human designed networks and a set of search algorithms. we achieved an accuracy of $96\%$  on the validation-set of CIFAR10 dataset.
\begin{table}[H]
    \centering
    \tiny
    \caption{Compare CIFAR10 Validation Accuracy}
    \label{tab:cifar10_acc}
    \begin{tabular}{|l|c|c|c|c|r} 
      \hline
      Model  & GPUs& days & Error\\
      \hline
      Network in Network \cite{lin2013network} & -& -  & 8.81 \\
      Deeply Supervised Net\cite{lee2015deeply} & -& -  & 7.97 \\
      Highway Network \cite{srivastava2015highway} & - & -  & 7.72 \\
      ResNet \cite{He_2016_CVPR} & -&-  & 6.61\\
      DenseNet-BC \cite{iandola2014densenet} & -&-  & 3.46\\
      DenseNet + Shake-Shake \cite{gastaldi2017shake} & -&-  & 2.86\\
      \hline
      \hline
      NAS\cite{zoph2016neural} & 800& 21-28 & 4.47 \\
      NAS + more filters \cite{zoph2016neural}& 800& 21-28  & 3.65  \\
      Hierarchical NAS & 200& 1.5  & 3.63 \\
      Micro NAS + Q-Learning& 32& 3  & 3.60  \\
      Progressive NAS \cite{liu2018progressive}& 100 & 1.5  & 3.63  \\
      NASNet-A \cite{zoph2018learning} & 450& 3-4 & 3.41 \\
      NASNet-A + CutOut \cite{zoph2018learning}& 450& 3-4  & 2.65 \\
      ENAS \cite{pham2018efficient}& 1& 0.45 & 3.54  \\
      ENAS + CutOut \cite{pham2018efficient}& 1& 0.45  & 2.89  \\
      DARTS (first order) + cutout\cite{liu2018darts} & 1 & 1.5 & 2.94 \\
      \hline
      \hline
      GeneticNAS + CutOut & 1& 0.4 & 4.00 \\
      \hline
    \end{tabular}
\end{table}
\subsection{CIFAR100 Result}
We performed the same search on CIFAR100. A final training stage with the same hyperparameter as used on CIFAR10 was performed. except the number of channels in the search which increased to $N_c=48$ and the batch size reduced to $64$ in-order to fit into the GPU memory. This result took twice the run time compared to the CIFAR10 dataset. 
\begin{figure}[H]
	\includegraphics[scale=0.4]{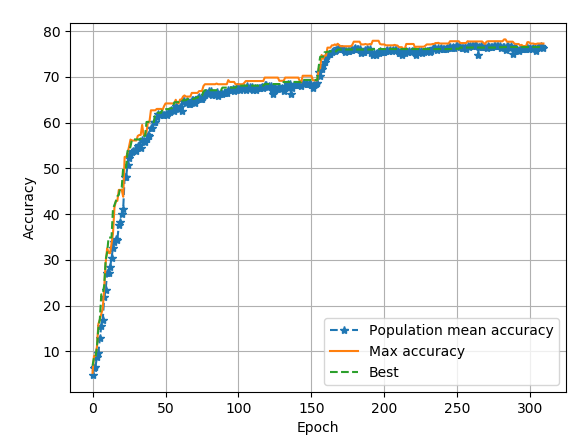}
	\centering
	\caption{Comparing the accuracy of the population on a subset of the validation-set vs the accuracy of the best model on the entire validation-set, performed on CIFAR100 dataset}
	\label{fig:pop_cifar100}
\end{figure}
Figure \ref{fig:pop_cifar100} presents a small difference between the max accuracy of the population that is measured on a subset of the validation-set and the best accuracy which is measured on the entire validation-set. The small difference indicates that measurement on the subset of the validation set is an acceptable prediction for the accuracy over the entire validation-set.\\
Once we have found a architecture we run a final training stage with the same parameter used in CIFAR10 dataset. The result of the final training is $80.1\%$. 
\section{Conclusion}\label{discussion}
In this work, we have studied the use of genetic algorithms as a search method of neural network architectures. Additionally we used the weight sharing in order to reduce the search time. The ablation study focuses on exploration vs exploitation trade-off of the genetic algorithm. This has shown a large affect on the validation accuracy which can be addressed as regularization. We have shown that a GeneticNAS works well on both CIFAR10 and CIFAR100 datasets and achieved a final accuracy of $96\%$ and $80.1\%$ . GeneticNAS’s key contribution is the integration of genetic search method into the training process that applies architecture learning during training.
\bibliographystyle{IEEEbib}
\bibliography{ref}

\begin{thebibliography}{10}

\bibitem{zoph2016neural}
Barret Zoph and Quoc~V Le,
\newblock ``Neural architecture search with reinforcement learning,''
\newblock {\em arXiv preprint arXiv:1611.01578}, 2016.

\bibitem{zoph2018learning}
Barret Zoph, Vijay Vasudevan, Jonathon Shlens, and Quoc~V Le,
\newblock ``Learning transferable architectures for scalable image
  recognition,''
\newblock in {\em Proceedings of the IEEE conference on computer vision and
  pattern recognition}, 2018, pp. 8697--8710.

\bibitem{cai2018efficient}
Han Cai, Tianyao Chen, Weinan Zhang, Yong Yu, and Jun Wang,
\newblock ``Efficient architecture search by network transformation,''
\newblock AAAI, 2018.

\bibitem{liu2018progressive}
Chenxi Liu, Barret Zoph, Maxim Neumann, Jonathon Shlens, Wei Hua, Li-Jia Li,
  Li~Fei-Fei, Alan Yuille, Jonathan Huang, and Kevin Murphy,
\newblock ``Progressive neural architecture search,''
\newblock in {\em Proceedings of the European Conference on Computer Vision
  (ECCV)}, 2018, pp. 19--34.

\bibitem{liu2018darts}
Hanxiao Liu, Karen Simonyan, and Yiming Yang,
\newblock ``Darts: Differentiable architecture search,''
\newblock {\em arXiv preprint arXiv:1806.09055}, 2018.

\bibitem{pham2018efficient}
Hieu Pham, Melody~Y Guan, Barret Zoph, Quoc~V Le, and Jeff Dean,
\newblock ``Efficient neural architecture search via parameter sharing,''
\newblock {\em arXiv preprint arXiv:1802.03268}, 2018.

\bibitem{8237416}
L.~Xie and A.~Yuille,
\newblock ``Genetic cnn,''
\newblock in {\em 2017 IEEE International Conference on Computer Vision
  (ICCV)}, Oct 2017, pp. 1388--1397.

\bibitem{liu2017hierarchical}
Hanxiao Liu, Karen Simonyan, Oriol Vinyals, Chrisantha Fernando, and Koray
  Kavukcuoglu,
\newblock ``Hierarchical representations for efficient architecture search,''
\newblock {\em arXiv preprint arXiv:1711.00436}, 2017.

\bibitem{real2018regularized}
Esteban Real, Alok Aggarwal, Yanping Huang, and Quoc~V Le,
\newblock ``Regularized evolution for image classifier architecture search,''
\newblock {\em arXiv preprint arXiv:1802.01548}, 2018.

\bibitem{krizhevsky2009learning}
Alex Krizhevsky and Geoffrey Hinton,
\newblock ``Learning multiple layers of features from tiny images,''
\newblock Tech. {R}ep., Citeseer, 2009.

\bibitem{He_2016_CVPR}
Kaiming He, Xiangyu Zhang, Shaoqing Ren, and Jian Sun,
\newblock ``Deep residual learning for image recognition,''
\newblock in {\em The IEEE Conference on Computer Vision and Pattern
  Recognition (CVPR)}, June 2016.

\bibitem{devries2017improved}
Terrance DeVries and Graham~W Taylor,
\newblock ``Improved regularization of convolutional neural networks with
  cutout,''
\newblock {\em arXiv preprint arXiv:1708.04552}, 2017.

\bibitem{xie2017aggregated}
Saining Xie, Ross Girshick, Piotr Doll{\'a}r, Zhuowen Tu, and Kaiming He,
\newblock ``Aggregated residual transformations for deep neural networks,''
\newblock in {\em Computer Vision and Pattern Recognition (CVPR), 2017 IEEE
  Conference on}. IEEE, 2017, pp. 5987--5995.

\bibitem{howard2017mobilenets}
Andrew~G Howard, Menglong Zhu, Bo~Chen, Dmitry Kalenichenko, Weijun Wang,
  Tobias Weyand, Marco Andreetto, and Hartwig Adam,
\newblock ``Mobilenets: Efficient convolutional neural networks for mobile
  vision applications,''
\newblock {\em arXiv preprint arXiv:1704.04861}, 2017.

\bibitem{szegedy2017inception}
Christian Szegedy, Sergey Ioffe, Vincent Vanhoucke, and Alexander~A Alemi,
\newblock ``Inception-v4, inception-resnet and the impact of residual
  connections on learning.,''
\newblock in {\em AAAI}, 2017, vol.~4, p.~12.

\bibitem{sandler2018mobilenetv2}
Mark Sandler, Andrew Howard, Menglong Zhu, Andrey Zhmoginov, and Liang-Chieh
  Chen,
\newblock ``Mobilenetv2: Inverted residuals and linear bottlenecks,''
\newblock {\em arXiv preprint arXiv:1801.04381}, 2018.

\bibitem{ioffe2015batch}
Sergey Ioffe and Christian Szegedy,
\newblock ``Batch normalization: Accelerating deep network training by reducing
  internal covariate shift,''
\newblock {\em arXiv preprint arXiv:1502.03167}, 2015.

\bibitem{hu2018squeeze}
Jie Hu, Li~Shen, and Gang Sun,
\newblock ``Squeeze-and-excitation networks,''
\newblock in {\em Proceedings of the IEEE Conference on Computer Vision and
  Pattern Recognition}, 2018, pp. 7132--7141.

\bibitem{burke2005search}
Edmund~K Burke, Graham Kendall, et~al.,
\newblock {\em Search methodologies},
\newblock Springer, 2005.

\bibitem{srivastava2014dropout}
Nitish Srivastava, Geoffrey Hinton, Alex Krizhevsky, Ilya Sutskever, and Ruslan
  Salakhutdinov,
\newblock ``Dropout: a simple way to prevent neural networks from
  overfitting,''
\newblock {\em The Journal of Machine Learning Research}, vol. 15, no. 1, pp.
  1929--1958, 2014.

\bibitem{sutskever2013importance}
Ilya Sutskever, James Martens, George Dahl, and Geoffrey Hinton,
\newblock ``On the importance of initialization and momentum in deep
  learning,''
\newblock in {\em International conference on machine learning}, 2013, pp.
  1139--1147.

\bibitem{gastaldi2017shake}
Xavier Gastaldi,
\newblock ``Shake-shake regularization,''
\newblock {\em arXiv preprint arXiv:1705.07485}, 2017.

\bibitem{larsson2016fractalnet}
Gustav Larsson, Michael Maire, and Gregory Shakhnarovich,
\newblock ``Fractalnet: Ultra-deep neural networks without residuals,''
\newblock {\em arXiv preprint arXiv:1605.07648}, 2016.

\bibitem{lin2013network}
Min Lin, Qiang Chen, and Shuicheng Yan,
\newblock ``Network in network,''
\newblock {\em arXiv preprint arXiv:1312.4400}, 2013.

\bibitem{lee2015deeply}
Chen-Yu Lee, Saining Xie, Patrick Gallagher, Zhengyou Zhang, and Zhuowen Tu,
\newblock ``Deeply-supervised nets,''
\newblock in {\em Artificial Intelligence and Statistics}, 2015, pp. 562--570.

\bibitem{srivastava2015highway}
Rupesh~Kumar Srivastava, Klaus Greff, and J{\"u}rgen Schmidhuber,
\newblock ``Highway networks,''
\newblock {\em arXiv preprint arXiv:1505.00387}, 2015.

\bibitem{iandola2014densenet}
Forrest Iandola, Matt Moskewicz, Sergey Karayev, Ross Girshick, Trevor Darrell,
  and Kurt Keutzer,
\newblock ``Densenet: Implementing efficient convnet descriptor pyramids,''
\newblock {\em arXiv preprint arXiv:1404.1869}, 2014.

\end{thebibliography}
\onecolumn
\section{Appendix}
\subsection{Code and Implantation}
The code and implantation of GeneticNAS which used for running the models is available online in the flowing github repository \href{https://github.com/haihabi/GeneticNAS}{https://github.com/haihabi/GeneticNAS}.
\subsection{Search Result}
\begin{figure}[H]
	\includegraphics[scale=0.3]{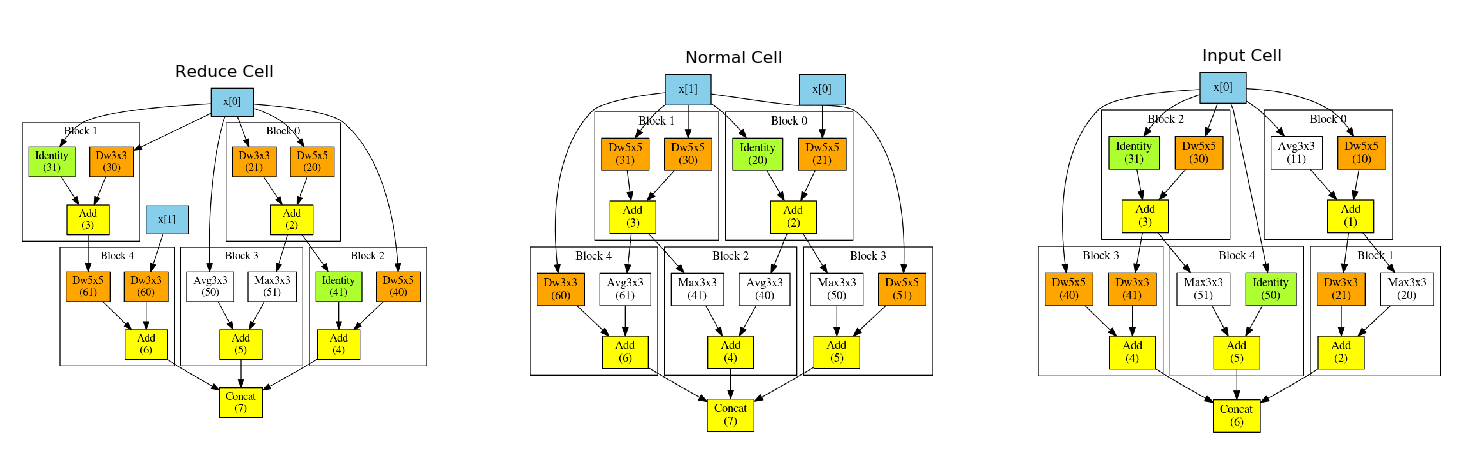}
	\centering
	\caption{Reduce Cell, Normal Cell and Input Cell Search Result on CIFAR100 dataset}
	\label{fig:search_res_cifar100}
\end{figure}
\end{document}